# Quantifier Scope and Constituency

Jong C. Park
Computer and Information Science
University of Pennsylvania
200 South 33rd Street, Philadelphia, PA 19104-6389, USA
park@linc.cis.upenn.edu

## Abstract

Traditional approaches to quantifier scope typically need stipulation to exclude readings that are unavailable to human understanders. This paper shows that quantifier scope phenomena can be precisely characterized by a semantic representation constrained by surface constituency, if the distinction between referential and quantificational NPs is properly observed. A CCG implementation is described and compared to other approaches.

## 1 Introduction

It is generally assumed that sentences with multiple quantified NPs are to be interpreted by one or more unambiguous logical forms in which the scope of traditional logical quantifiers determines the reading or readings. There are two problems with this assumption: (a) without further stipulation there is a tendency to allow *too many* readings and (b) there is considerable confusion as to *how many* readings should be allowed arising from contamination of the semantics of many NL quantifiers by referentiality.

There are two well-known techniques for redistributing quantifiers in quantification structures: quantifying-in (Montague, 1974; Cooper, 1983; Keller, 1988; Carpenter, 1994) and quantifier raising (May, 1985). The former provides a compositional way of putting possibly embedded quantifiers to the scope-taking positions, and the latter utilizes a syntactic movement operation at the level of semantics for quantifier placement. There are also approaches that put more emphasis on utilizing contextual information in restricting the generation of semantic forms by choosing a scope-neutral representation augmented with ordering constraints to capture linguistic judgments (Webber, 1979; Kamp, 1981; Heim, 1983; Poesio, 1991; Reyle, 1993). And there are computational approaches that screen unavailable and/or redundant semantic forms (Hobbs & Shieber, 1987; Moran, 1988; Vestre, 1991). This paper will show that these approaches allow unavailable readings, and thereby miss an important generalization concerning the readings that actually are available.

This paper examines English constructions that allow multiple occurrences of quantified NPs: NP modifications, transitive or ditransitive verbs, *that* complements, and coordinate structures. Based on a critical analysis of readings that are available from these data, the claim is that scope phenomena can be characterized by a combination of syntactic surface adjacency and semantic function-argument relationship. This characterization will draw upon the old distinction between referential and quantificational NP-semantics (Fodor & Sag, 1982). We choose to use Combinatory Categorial Grammar to show how surface adjacency affects semantic function-argument relationship, since CCG has the flexibility of composing almost any pair of adjacent constituents with a precise notion of syntactic grammaticality (Steedman, 1990; 1993).[1]

The rest of the paper is organized as follows. First, we discuss in §2 how traditional techniques address availability of readings and note some residual problems. Then we give a brief analysis of available readings (§3), a generalization of the analysis (§4), and finally describe a computational implementation in Prolog (§5).

## 2 Traditional Approaches

All three paradigms of grammar formalisms introduced earlier share similar linguistic judgments for their grammaticality analyses. This section examines quantifying-in to show (a) that quantifying-in is a powerful device that allows referential NP-interpretations and (b) that quantifying-in is not sufficiently restricted to account for the available readings for quantificational NP-interpretations.

Quantifying-in is a technique originally introduced to produce appropriate semantic forms for *de re* in-terpretations of NPs inside opaque operators (Montague, 1974). For example, (a) below has two readings, *de re* and *de dicto*, depending on the relativity of the existence of such an individual. They are roughly interpretable as (b) and (c).[2]

---

[1] For instance, the result would transfer to Synchronous Tree Adjoining Grammar (Shieber & Schabes, 1990) without much change.

(1) (a) John believes that a Republican will win.
    (b) `∃r.repub(r) ∧ bel(john, will(win(r)))`
    (c) `bel(john, ∃r.repub(r) ∧ will(win(r)))`

(b) has a binder ∃ that is *quantifying* a variable `r` *in*side an opaque operator `bel`, hence the name for the technique. (c) does not have such an intervening operator. Although it is beyond the scope of the present paper to discuss further details of intensionality, it is clear that *de re* interpretations of NPs are strongly related to referential NP-semantics, in the sense that the *de re* reading of (a) is about a referred individual and not about an arbitrary such individual. Quantifying-in is designed to make *any* (possibly embedded) NP take the matrix scope, by leaving a scoped variable in the argument position of the original NP. This would be acceptable for referential NP-semantics.

Montague also proposed to capture purely extensional scope ambiguities using quantifying-in. For example, wide scope reading of *a woman* in (a) below is accounted for by quantifying-in (with a meaning postulate), patterned after one for (b).

(2) (a) Every man loves a woman.
    (b) Every man seeks a white unicorn.

His suggestion is adopted with various subsequent revisions cited earlier. Since any NP, referential or quantificational, requires quantifying-in to outscope another, quantifying-in consequently confounds referential and quantificational NP-semantics. This causes a problem when there is a distributional difference between referential NPs and non-referential NPs, as Fodor & Sag (1982) have argued, a view which has been followed by the approaches to dynamic interpretation of indefinite NPs cited earlier. It seems hard to reconcile quantifying-in with these observations.

## 3 Availability of Readings

This section proposes a way of sharpening our intuition on available readings and re-examines traditional linguistic judgments on grammatical readings.

While there are undoubted differences in degree of availability among readings dependent upon semantics or discourse preference (Bunt, 1985; Moran, 1988), we will focus on <u>all-or-none</u> structural possibilities afforded by competence grammar.[3]

Consider the following unambiguous quantification structure in a generalized quantifier format (hereafter GQ, Barwise & Cooper, 1981), where **quantifier** *outscopes* any quantifiers that may occur in either **restriction** or **body**.

(3) `quantifier(variable, restriction, body)`

Logical forms as notated this way make explicit the functional dependency between the denotations of two ordered quantificational NPs. For example, consider (4) (a) (Partee, 1975). (b) shows one way of representing it in a GQ format.

(4) (a) Three Frenchmen visited five Russians.
    (b) `three(f, frenchmen(f), five(r, russians(r), visited(f,r)))`

We can always argue, by enriching the notation, that (4) (b) represents at least four different readings, depending on the particular sense of each involved NP, i.e., group- vs individual-denoting. In every such reading, however, the truth of (4) (b) depends upon finding appropriate individuals (or the group) for `f` such that *each* of those individuals (or the group itself) gets associated with appropriate individuals (or a group of individuals) for `r` via the relation **visited**.[4] Notice that there is always a *functional dependency* of individuals denoted by `r` upon individuals denoted by `f`. We claim that this explicit functional dependency can be utilized to test availability of readings.[5]

First, consider the following sentences without coordination.

(5) (a) Two representatives of three companies saw most samples.
    (b) Every dealer shows most customers at most three cars.
    (c) Most boys think that every man danced with two women.

(a) has three quantifiers, and there are 6 different ways of ordering them. Hobbs & Shieber (1987) show that among these, the reading in which *two representatives* outscopes *most samples* which in turn outscopes *three companies* is not available from the sentence. They attribute the reason to the logical structure of English as in (3), as it is considered unable to afford an unbound variable, a constraint known as the unbound variable constraint (UVC).[6]

---

[2] In this simplistic notation, we gloss over tense analysis, among others.

[3] Moran's preference-based algorithm treats certain readings as "highly unpreferred," effectively making them structurally unavailable, from those possible scopings generated by a scheme similar to Hobbs & Shieber (1887). We claim that competence grammar makes even fewer readings available in the first place.

[4] Without losing generality, therefore, we will consider only individual-denoting NPs in this paper.

[5] Singular NPs such as *a company* are not helpful to this task since their denotations do not involve multiple individuals which explicitly induce this functional dependency.

[6] The reading would be represented as follows, which has the first occurrence of the variable `c` left unbound.

We should note, however, that there is one reading among the remaining five that the UVC allows which in fact does not appear to be available. This is the one in which *three companies* outscopes *most samples* which in turn outscopes *two representatives* (cf. Horn (1972), Fodor (1982)).[7] This suggests that the UVC may not be the only principle under which Hobbs & Shieber's reading is excluded.[8] The other four readings of (a) are self-evidently available. If we generalize over available readings, they are only those that have no quantifiers which intercalate over NP boundaries.[9]

(5) (b) has three quantifiers too, but unlike (5) (a), all the six ways of ordering the quantifiers are available. (5) (c) has only four available readings, where *most boys* does not intercalate *every man* and *two women*.[10]

Consider now sentences including coordination.

(6) (a) Every girl admired, but most boys detested, one of the saxophonists.
    (b) Most boys think that every man danced with, but doubt that a few boys talked to, more than two women.

As Geach (1970) pointed out, (a) has only two grammatical readings, though it has three quantifiers. In reading 1, the same saxophonist was admired and detested at the same time. In reading 2, every girl admired an arbitrary saxophonist and most boys also detested an arbitrary saxophonist. In particular, missing readings include the one in which every girl admired the same saxophonist and most

```
two(r, rep(r) & of(r,c), most(s, samp(s),
three(c, comp(c), saw(r,s))))
```

[7]To paraphrase this impossible reading, it is true of a situation under which there were three companies such that there were four samples for each such company such that each such sample was seen by two representatives of that company. Crucially, samples seen by representatives of different companies were not necessarily the same.

[8]This should not be taken as denying the reality of the UVC itself. For example, as one of the referees pointed out, the UVC is required to explain why, in (a) below, *every professor* must outscope *a friend* so as to bind the pronoun *his*.

(a) Most students talked to a friend of every professor about his work.

[9]One can replace *most samples* with other complex NP such as *most samples of at least five products* to see this. Certain sentences that apparently escape this generalization will be discussed in the next section.

[10]To see why they are available, it is enough to see that (a) and (b) below have two readings each.

(a) John thinks that every man danced with two women.
(b) Most boys think that Bill danced with two women.

boys detested the same but another saxophonist. (6) (b) also has only two grammatical readings. In one, *most boys* outscopes *every man* and *a few boys* which together outscope *more than two women*. In the other, *more than two women* outscopes *every man* and *a few boys*, which together outscope *most boys*.

## 4 An Account of Availability

This section proposes a generalization at the level of semantics for the phenomena described earlier and considers its apparent counterexamples.

Consider a language $\mathcal{L}$ for natural language semantics that explicitly represents function-argument relationships (Jackendoff, 1972). Suppose that in $\mathcal{L}$ the semantic form of a quantified NP is a syntactic argument of the semantic form of a verb or a preposition. (7) through (10) below show well-formed expressions in $\mathcal{L}$.[11]

(7) `visited(five(russian),three(frenchman))`

(8) `saw(most(samp),of(three(comp),two(rep)))`

(9) `show(three(car),most(cstmr),every(dlr))`

(10) `think(^danced(two(woman),every(man)), most(boy))`

For instance, `of` has two arguments `three(comp)` and `two(rep)`, and `show` has three arguments.

$\mathcal{L}$ gives rise to a natural generalization of available readings as summarized below.[12]

(11) For a function with $n$ arguments, there are $n!$ ways of successively providing all the arguments to the function.

This generalization captures the earlier observations about availability of readings. (7), for (4) (a), has two (2!) readings, as `visited` has two arguments. (8) is an abstraction for four (2!×2!) readings, as both `of` and `saw` have two arguments each. (9) is an abstraction for six (3!) readings, as `show` has three arguments. Likewise, (10) is an abstraction for four readings.

Coordination gives an interesting constraint on availability of readings. Geach's observation that (6) (a) has two readings suggests that the scope of the object must be determined *before* it reduces with the coordinate fragment. Suppose that the non-standard constituent for one of the conjuncts in (6) (a) has a semantic representation shown below.

(12) $\lambda x$ `admired(`$x$`,every(girl))`

Geach's observation implies that (12) is ambiguous, so that `every(girl)` can still take wide (or narrow)

[11]The up-operator ^ in (10) takes a term of type t to a term of type e, but a further description of $\mathcal{L}$ is not relevant to the present discussion.

[12]Nam (1991)'s work is based on a related observation, though he does not make use of the distinction between referential and quantificational NP-semantics.

scope with respect to the unknown argument. A theory of CCG will be described in the next section to show how to derive scoped logical forms for available readings only.

But first we must consider some apparent counterexamples to the generalization.

(13) (a) Three hunters shot at five tigers.
     (b) Every representative of a company saw most samples.

The obvious reading for (a) is called conjunctive or cumulative (Partee, 1975; Webber 1979). In this reading, there are three hunters and five tigers such that shooting events happened between the two parties. Here, arguments are not presented *in succession* to their function, contrary to the present generalization. Notice, however, that the reading must have two (or more) referential NPs (Higginbotham, 1987).[13] The question is whether our theory should predict this possibility as well. For a precise notion of availability, we claim that we must appeal to the distinction between referential and quantificational NP-semantics, since almost *any* referential NP can have the appearance of taking the matrix scope, without affecting the rest of scope phenomena. A related example is (b), where in one reading a referential NP *a company* arguably outscopes *most samples* which in turn outscopes *every representative* (Hobbs & Shieber, 1987). As we have pointed out earlier, the reading does not generalize to quantified NPs in general.

(14) (a) Some student will investigate two dialects of every language.
     (b) Some student will investigate two dialects of, and collect all interesting examples of coordination in, every language.
     (c) * Two representative of at least three companies touched, but of few universities saw, most samples.

(a) has a reading in which *every language* outscopes *some student* which in turn outscopes *two dialects* (May, 1985). In a sense, this has intercalating NP quantifiers, an apparent problem to our generalization. However, the grammaticality of (b) opens up the possibility that the two conjuncts can be represented grammatically as functions of arity two, similar to normal transitive verbs. Notice that the generalization is not at work for the fragment *of at least three companies touched* in (c), since the conjunct is syntactically ungrammatical. At the end of next section, we show how these finer distinctions are made under the CCG framework (See discussion of Figure 5).

---
[13]For example, (a) below lacks such a reading.

(a) Several men danced with few women.

## 5 A CCG Implementation

This section describes a CCG approach to deriving scoped logical forms so that they range over only grammatical readings.

We will not discuss details of how CCG characterizes natural language syntactically, and refer the interested reader to Steedman (1993). CCGs make use of a limited set of combinators, type raising (**T**), function composition (**B**), and function substitution (**S**), with directionality of combination for syntactic grammaticality. For the examples in this paper, we only need type raising and function composition, along with function application. The following shows rules of derivation that we use. Each rule is associated with a label, such as > or <B etc, shown at the end.

(15) (a) X/Y    Y      => X            (>)
     (b) Y      X\Y    => X            (<)
     (c) X/Y    Y/Z    => X/Z          (>B)
     (d) Y\Z    X\Y    => X\Z          (<B)
     (e)        np     => T/(T\np)     (>T)
     (f)        np     => T\(T/np)     (<T)

The mapping from syntax to semantics is usually defined in two different ways. One is to use elementary categories, such as np or s, in encoding both syntactic types and logical forms (Jowsey, 1990; Steedman, 1990; Park, 1992). The other is to associate the entire lexical category with a higher-order expression (Kulick, 1995). In this paper, we take the former alternative to describe a first-order rendering of CCG.

Some lexical entries for *every* are shown below.

(16) (s:q-every(X,N,S)/(s:S\np:X))/n:X^N
(17) (s:S/(s:S\np:s-every(N)))/n:N

The information (s/(s\np))/n encodes the syntactic fact that *every* is a constituent which, when a constituent of category n is provided on its right, returns a constituent of category s/(s\np). q-every(X,N,S) is a term for scoped logical forms. We are using different lexical items, for instance q-every and s-every for *every*, in order to signify their semantic differences.[14] These lexical entries are just two instances of a general schema for type-raised categories of quantifiers shown below, where T is an arbitrary category.

(18) (T/(T\np))/n and (T\(T/np))/n

And the semantic part of (16) and (17) is first-order encoding of (19) (a) and (b), respectively.[15]

---
[14]q-every represents *every* as a quantifier, and s-every, as a set denoting property. We will use s-every(X^man(X)) and its $\eta$-reduced equivalent s-every(man) interchangeably.

[15]s-quantifier(noun) denotes an arbitrary set $N$ of individuals $d$ such that $d$ has the property noun and that the cardinality of $N$ is determined by quantifier (and

(19) (a) $\lambda n.\lambda P.\forall x \in \text{s-every}(n).P(x)$
     (b) $\lambda n.\lambda P.P(\text{s-every}(n))$

(a) encodes wide scope type raising and (b), narrow.

With standard entries for verbs as in (20), logical forms such as (21) and (22) are possible.

(20) `saw :- (s:saw(X,Y)\np:X)/np:Y`
(21) `q-two(X,rep(X),saw(X,s-four(samp)))`
(22) `q-two(X,rep(X),q-four(Y,samp(Y),saw(X,Y)))`

Figure 1 shows different ways of deriving scoped logical forms. In (a), `n:X^N` *unifies* with `n:X^girl(X)`, so that `N` gets the value `girl(X)`. This value of `N` is transferred to the expression `s:every(X,N,S)` by partial execution (Pereira & Shieber, 1987; Steedman, 1990; Park, 1992). (a) shows a derivation for a reading in which object NP takes wide scope and (b) shows a derivation for a reading in which subject NP takes wide scope. There are also other derivations.

Figure 2 shows logical forms that can be derived in the present framework from Geach's sentence. Notice that the conjunction forces subject NP to be first composed with the verb, so that subject NP must be type-raised *and* be combined with the semantics of the transitive verb. As noted earlier, the two categories for the object still make both scope possibilities available, as desired. The following category is used for *but*.

(23) `((s:and(P,Q)/np:X)\(s:P/np:X))/(s:Q/np:X)`

Readings that involve intercalating quantifiers, such as the one where *every girl* outscopes *one saxophonist*, which in turn outscopes *most boys*, are correctly excluded.

Figure 3 shows two different derivations of logical forms for the complex NP *two representatives of three companies*. (a) shows a derivation for a reading in which the modifying NP takes wide scope and (b) shows the other case. In combination with derivations involving transitive verbs with subject and object NPs, such as ones in Figure 1, this correctly accounts for four grammatical readings for (5) (a).[16]

Figure 4 shows a derivation for a reading, among six, in which *most customers* outscopes *every dealer* which in turn outscopes *three cars*. Some of these readings become unavailable when the sentence contains coordinate structure, such as one below.

(24) Every dealer shows most customers (at most) three cars but most mechanics every car.

---

noun). We conjecture that this can also be made to capture several related NP-semantics, such as collective NP-semantics and/or referential NP-semantics, though we can not discuss further details here.

[16]As we can see in Figure 3 (a) & (b), there is no way quantifiers inside `S` can be placed *between* the two quantifiers `two` & `three`, correctly excluding the other two readings.

In particular, (24) does not have those two readings in which *every dealer* intercalates *most customers* and *three cars*. This is exactly predicted by the present CCG framework, extending Geach's observation regarding (6) (a), since the coordination forces the two NPs, *most customers* and *three cars*, to be composed first (Dowty, 1988; Steedman 1990; Park 1992). (25) through (27) show one such derivation, which results in readings where *three cars* outscopes *most customers* but *every dealer* must take either wide or narrow scope with respect to both *most customers* and *three cars*.

(25)
```
    most customers
    -----------------------------------
    ((s:q-most(Z,cstmr(Z),S)\np:X)/np:Y)
    \(((s:S\np:X)/np:Y)/np:Z)
```

(26)
```
    three cars
    ----------------------------
    (s:q-three(Y,car(Y),S)\np:X)
    \((s:S\np:X)/np:Y)
```

(27)
```
    most customers              three cars
    --------------              ----------
     see above                  see above
    ------------------------------------------<B
    (s:q-three(Y,car(Y),q-most(Z,cstmr(Z),S))
    \np:X)\(((s:S\np:X)/np:Y)/np:Z)
```

Figure 5 shows the relevant derivation for the fragment *investigate two dialects of* discussed at end of previous section. It is a conjoinable constituent, but since there is no way of using type-raised category for *two* for a successful derivation, *two dialects* can not outscope any other NPs, such as subject NP or the modifying NP (Steedman, 1992). This correctly accounts for our intuition that (14) (a) has an apparently intercalating reading and that (14) (b) has only two readings. However, there is no similar derivation for the fragment *of three companies touched*, as shown below.

(28)
```
      of         three companies    touched
    --------     ---------------    ---------
    (n\n)/np       T\(T/np)         (s\np)/np
    ------------------------------<
         n\n (with T = n\n)
    ------------------------------------*
```

## 6 Concluding Remarks

We have shown that the range of grammatical readings allowed by sentences with multiple quantified NPs can be characterized by abstraction at function-argument structure constrained by syntactic adjacency. This result is in principle available to other paradigms that invoke operations like QR at LF or type-lifting, which are essentially equivalent to abstraction. The advantage of CCG's very free notion of surface structure is that it ties abstraction or the equivalent as closely as possible to derivation. Apparent counterexamples to the generalization can be

```
(a)    every                      girl              admired                     one saxophonist
       ------------------         -----------       ---------------------       ------------------------------
       s:q-every(X,N,S)           n:X^girl(X)       (s:admired(X,Y)\np:X)       s:q-one(Y,sax(Y),S)\(s:S/np:Y)
       /(s:S\np:X)/n:X^N                            /np:Y
       ------------------------------------->
       s:q-every(X,girl(X),S)/(s:S\np:X)
       ---------------------------------------------------------->B
               s:q-every(X,girl(X),admired(X,Y))/np:Y
       ------------------------------------------------------------------------------------------------<
               s:q-one(Y,sax(Y),q-every(X,girl(X,admired(X,Y))))

(b)    every girl                                          admired                    one saxophonist
       ---------------------------------                   ---------------------      ------------------------
       s:q-every(X,girl(X),S)/(s:S\np:X)                   (s:admired(X,Y)\np:X)      s:S\(s:S/np:s-one(sax))
                                                           /np:Y
       --------------------------------------------------->B
              s:q-every(X,girl(X),admired(X,Y))/np:Y
       ----------------------------------------------------------------------------------<
                      s:q-every(X,girl(X),admired(X,s-one(sax)))
```

Figure 1: *Every girl admired one saxophonist*: Two sample derivations

```
(a)    every girl admired                                but    most boys detested    one saxophonist
       ------------------------------------              ------------------           ------------------------
       s:q-every(X,girl(X),admired(X,Y))/np:Y            -------------------->        s:S\(s:S/np:s-one(sax))
       -----------------------------------------------------------------<
       s:and(q-every(X,girl(X),admired(X,Y)),q-most(X,boy(X),detested(X,Y)))/np:Y
       ---------------------------------------------------------------------------------------------------<
          s:and(q-every(X,girl(X),admired(X,s-one(sax))),q-most(X,boy(X),detested(X,s-one(sax))))

(b)    every girl admired                          but    most boys detested    one saxophonist
       ------------------------------              ------------------           ------------------------------
       s:admired(s-every(girl),Y)/np:Y             -------------------->        s:q-one(Y,sax(Y),S)\(s:S/np:Y)
       -------------------------------------------------------<
       s:and(admired(s-every(girl),Y),detested(s-most(boy),Y))/np:Y
       -----------------------------------------------------------------------------------<
              s:q-one(Y,sax(Y),and(admired(s-every(girl),Y),detested(s-most(boy),Y)))
```

Figure 2: *Every girl admired, but most boys detested, one saxophonist*: Two sample derivations

```
(a)    two                     representatives      of        three companies
       ------------------      --------------------------      ---------------------------------------
       (s:q-two(X,N,S)         n:X^and(rep(X),of(X,Y))/np:Y    (s:q-three(C,comp(C),S2)/(s:S1\np:X))
        /(s:S\np:X))/n:X^N                                     \((s:S2/(s:S1\np:X))/np:C)
       ------------------------------------------------->B
       (s:q-two(X,and(rep(X),of(X,Y)),S)/(s:S\np:X))/np:Y
       ----------------------------------------------------------------------------------------<
              s:q-three(C,comp(C),q-two(X,and(rep(X),of(X,C)),S))/(s:S\np:X)

(b)    two                     representatives     of         three companies
       ------------------      --------------------------     -------------------------------
       (s:q-two(X,N,S)         n:X^and(rep(X),of(X,Y))/np:Y   (s:S2/(s:S1\np:X))
       /(s:S\np:X))/n:X^N                                     \((s:S2/(s:S1\np:X))/np:s-three(comp))
       ------------------------------------------------->B
       (s:q-two(X,and(rep(X),of(X,Y)),S)/(s:S\np:X))/np:Y
       ----------------------------------------------------------------------------------------<
              s:q-two(X,and(rep(X),of(X,s-three(comp))),S)/(s:S\np:X)
```

Figure 3: *two representatives of three companies*: Two sample derivations

```
every dealer              shows                  most customers           three cars
------------              ---------------------  -----------------------  -------------------
s:q-every(X,dlr(X),S)     (s:show(X,Y,Z)\np:X)   (s:q-most(Y,cstmr(Y),S)  s:S\(s:S
/(s:S\np:X)               /np:Z/np:Y             /np:Z)\(s:S/np:Z)/np:Y   /np:s-three(car))
----------------------------------------->B
s:q-every(X,dlr(X),show(X,Y,Z))/np:Z/np:Y
---------------------------------------------------------------------<
       s:q-most(Y,cstmr(Y),q-every(X,dlr(X),show(X,Y,Z)))/np:Z
------------------------------------------------------------------------------------<
       s:q-most(Y,cstmr(Y),q-every(X,dlr(X),show(X,Y,s-three(car))))
```

Figure 4: *Every dealer shows most customers three cars*: One sample derivation

```
investigate                 two         dialects           of
------------------------    ----------  ---------------    --------------------------
(s:investigate(X,Y)\np:X)   np:s-two(N) n:N1/(n:N1         (n:Y^and(N,of(Y,Z))\n:Y^N)
/np:Y                       /n:N        \n:Y^dialect(Y))   /np:Z
                                        ------------------------------------------>B
                                        n:Y^and(dialect(Y),of(Y,Z))/np:Z
                            ---------------------------------------------------->B
                            np:s-two(Y^and(dialect(Y),of(Y,Z)))/np:Z
------------------------------------------------------------------------------->B
       (s:investigate(X,s-two(Y^and(dialect(Y),of(Y,Z)))\np:X)/np:Z
```

Figure 5: *investigate two dialects of*: One derivation

explained by the well-known distinction between referential and quantificational NP-semantics. An implementation of the theory for an English fragment has been written in Prolog, simulating the 2nd order properties.

There is a question of how the non-standard surface structures of CCG are compatible with well-known conditions on binding and control (including crossover). These conditions are typically stated on standard syntactic dominance relations, but these relations are no longer uniquely derivable once CCG allows non-standard surface structures. We can show, however, that by making use of the obliqueness hierarchy (cf. Jackendoff (1972) and much subsequent work) at the level of LF, rather than surface structure, it is possible to state such conditions (Steedman, 1993).

## Acknowledgements

Special thanks to Mark Steedman. Thanks also to Janet Fodor, Beryl Hoffman, Aravind Joshi, Nobo Komagata, Anthony Kroch, Michael Niv, Charles L. Ortiz, Jinah Park, Scott Prevost, Matthew Stone, Bonnie Webber, and Michael White for their help and criticism at various stages of the presented idea. Thanks are also due to the anonymous referees who made valuable suggestions to clarify the paper. Standard disclaimers apply. The work is supported in part by NSF grant nos. IRI91-17110, and CISE IIP, CDA 88-22719, DARPA grant no. N660001-94-C-6043, and ARO grant no. DAAH04-94-G0426.